\documentclass{acm_proc_article-sp}

\usepackage{paralist}

\begin{document}

\setlength{\pdfpagewidth}{8.5in}
\setlength{\pdfpageheight}{11in}

\title{Counterfactual Estimation and Optimization of Click Metrics for Search Engines
}
%
%
%
%
%

\numberofauthors{2} 
%
\author{
%
%
\alignauthor
Lihong Li$^1$ \hspace{15mm}
Shunbao Chen$^1$ \\
	   \affaddr{$^1$Microsoft Inc.}\\
       \affaddr{Redmond, WA 98052}\\
       \email{\{lihongli,shchen,ankurg\}@microsoft.com}
\alignauthor
Jim Kleban$^2$\titlenote{This work was done when J. Kleban was with Microsoft.} \hspace{15mm}
Ankur Gupta$^1$ \\
	   \affaddr{$^2$Facebook Inc.}\\
       \affaddr{Seattle, WA 98101}\\
       \email{jim.kleban@gmail.com}
}
\date{30 July 1999}

\newtheorem{definition}{Definition}
\newtheorem{lemma}{Lemma}
\newtheorem{prop}{Proposition}
\newtheorem{theorem}{Theorem}
\newtheorem{obs}{Observation}

\newcommand{\defref}[1]{Definition~\ref{#1}}
\newcommand{\lemref}[1]{Lemma~\ref{#1}}
\newcommand{\thmref}[1]{Theorem~\ref{#1}}
\newcommand{\secref}[1]{Section~\ref{#1}}
\newcommand{\secsref}[1]{Sections~\ref{#1}}
\newcommand{\obsref}[1]{Observation~\ref{#1}}
\newcommand{\eqnref}[1]{Equation~\ref{#1}}
\newcommand{\figref}[1]{Figure~\ref{#1}}
\newcommand{\algref}[1]{Algorithm~\ref{#1}}

\newcommand{\Xset}{\mathcal{X}}
\newcommand{\Aset}{\mathcal{A}}
\newcommand{\Dset}{\mathcal{D}}

\newcommand{\algfont}[1]{\textsc{#1}}
\newcommand{\aka}{\textit{a.k.a.}}
\newcommand{\cf}{\textit{c.f.}}
\newcommand{\eg}{\textit{e.g.}}
\newcommand{\etc}{\textit{etc.}}
\newcommand{\ie}{\textit{i.e.}}
\newcommand{\etal}{\textit{et al.}}
\newcommand{\iid}{i.i.d.}

\newcommand{\mdpmr}{MDP/R}
\newcommand{\pmin}{p_\mathrm{min}}
\newcommand{\vmax}{V_\mathrm{max}}
\newcommand{\amax}{A_\mathrm{max}}
\newcommand{\rmax}{\algfont{Rmax}}
\newcommand{\eee}{\algfont{E$^3$}}
\newcommand{\expert}{\mathcal{E}}
\newcommand{\defeq}{:=}
\newcommand{\Voff}{\hat{V}_\mathrm{offline}}
\newcommand{\Von}{\hat{V}_\mathrm{online}}

\newcommand{\abs}[1]{\left|#1\right|}
\newcommand{\bigO}[1]{\mathbb{O}\left(#1\right)}
\newcommand{\bigOt}[1]{\tilde{\mathbb{O}}\left(#1\right)}
\newcommand{\E}{\mathbf{E}}
\newcommand{\Var}{\mathbf{Var}}
\newcommand{\indfunc}[1]{\mathbb{I}\left(#1\right)}
\newcommand{\norm}[1]{\left\|#1\right\|}
\newcommand{\setcard}[1]{\left|#1\right|}

\newcommand{\namecite}[1]{\aunpcite{#1}~\yrcite{#1}}

\newcommand{\lihong}[1]{[[\textbf{Lihong:} \textit{#1}]]}

\graphicspath{{figs/}}

\maketitle
\begin{abstract}
Optimizing an interactive system against a predefined online metric is particularly challenging, when the metric is computed from user feedback such as clicks and payments.  The key challenge is the \emph{counterfactual} nature: in the case of Web search, any change to a component of the search engine may result in a different search result page for the same query, but we normally cannot infer reliably from search log how users would react to the new result page.  Consequently, it appears impossible to accurately estimate online metrics that depend on user feedback, unless the new engine is run to serve users and compared with a baseline in an A/B test.  This approach, while valid and successful, is unfortunately expensive and time-consuming.  In this paper, we propose to address this problem using causal inference techniques, under the contextual-bandit framework.  This approach effectively allows one to run  (potentially infinitely) many A/B tests \textit{offline} from search log, making it possible to estimate and optimize online metrics quickly and inexpensively.  Focusing on an important component in a commercial search engine, we show how these ideas can be instantiated and applied, and obtain very promising results that suggest the wide applicability of these techniques.
\end{abstract}




\section{Introduction}
\label{sec:intro}

The standard approach to evaluating ranking quality of a search engine is to evaluate its ranking results on a set of human-labeled examples and compute \emph{relevance metrics} like mean average precision (MAP)~\cite{Baezayates99Modern} and normalized discounted cumulative gain (NDCG)~\cite{Jarvelin02Cumulated}.  Such an approach has been highly successful at facilitating easy comparison and improvement of ranking functions (\eg, \cite{Burges05Learning,Trec,Zheng08General}).

However, such offline relevance metrics have a few limitations.  First, there can be a mismatch between users' actual information need and the relevance judgments of human labelers.  For example, for the query ``tom cruise,'' it is natural for a judge to give a high relevance score to the actor's official website, http://tomcruise.com.  However, search log from a commercial search engine suggests the opposite---users who issue that query are often more interested in news about the actor, not the official website.\footnote{The example is from Imed Zitouni.}  Second, in some applications like personalized search~\cite{Pitkow02Personalized} and recency search~\cite{Dong10Towards}, judges simply lack the information to provide sensible labels.  Third, user experience with a search engine relies on \emph{both} the ranking function and other modules like user interfaces.  Relevance labels for query-document pairs only reflect one aspect of a search engine's overall quality.  
Finally, an important factor in nowadays search engine is its monetization performance (from advertising), which cannot be easily judged by human labelers.

All the challenges above imply the strong need for considering user feedback in evaluating, and potentially optimizing, a search engine.  For example, user behavior like clicks is used to infer personalized relevance for evaluation purposes~\cite{Teevan10Potential}, and to compare two ranking systems by interleaving~\cite{Chapelle12Large}.

Unfortunately, metrics that depend on user feedback are hard to estimate \emph{offline}, due to their counterfactual nature.  For example, suppose we are interested in measuring the time-to-first-click metric.  When we change any part of the search engine, the final search engine result page (SERP) for a particular query may be different, and hence users' click behavior may change as well.  Based on search log, it is often challenging to infer what a user would have done for a SERP different from the one in the log.  Prediction errors of state-of-the-art user click models (\eg, \cite{Chapelle09Dynamic,Guo09Click} and their variants) are likely much larger than usual improvements of click-based metrics in nowadays commercial search engines that are already highly optimized.  Therefore, offline evaluation based on such user models may not always be reliable.

In practice, the common solution is to run a controlled experiment (\aka\ an A/B test).  Specifically, one randomly splits users into two statistically identical groups, known as \textit{control} and \textit{treatment}, respectively.  Users in the control group are served by a baseline search engine, while users in the treatment group by a modified engine (which often differs from the baseline in one component).  The experiment may last for days or weeks, at the end of which online metrics (like click-through rate and time to first click) of the two systems are calculated.  One then reaches a conclusion whether the modified engine is better than the baseline at a certain statistically significance level.

Controlled experiments have proved very successful in practice (\eg, \cite{Kohavi09Controlled}), allowing engineering and business decisions to be made in a data-driven manner.  However, these experiments usually require nontrivial engineering resources and are time-consuming, since the experiments are run on \emph{real} users, so significant efforts are needed to avoid surprises in the experiments.
Furthermore, when trying to optimize an online click metric, one often takes a guess-then-check approach: an easy-to-compute \emph{proxy} metric (like NDCG) is used offline to obtain a modified engine, which is hoped to do well later in the controlled experiment in terms of the click metric of real interest.  Due to its approximation nature, the proxy metric can be misleading in determining which modified system to run in the experiment.  Combined with the long turnaround time of A/B tests, this indirect optimization procedure can be rather inefficient.


In this paper, we advocate the use of causal inference techniques from statistics to perform \emph{unbiased offline evaluation of click metrics} for search engines.  Compared to A/B tests, offline evaluation allows multiple models to be evaluated on the \emph{same} search log, without the need to be run online.  Effectively, this technique makes it possible to run many A/B tests simultaneously, leading to substantial increase in experimentation agility, and to even optimize against the online metrics directly.  To the best of our knowledge, this work is the first to validate the possibility of offline evaluation in a live, commercial search engine, and to use such offline evaluation as a subroutine to do offline optimization.

The rest of the paper is organized as follows.  \secref{sec:bandit} describes the contextual bandit as a general framework to capture a number of interactive problems, including many in Web search.  \secref{sec:eval} describes the basic technical idea of offline evaluation, and discusses solutions to a few important issues that arise in practice.  \secref{sec:exp} gives details in a case study in a commercial search engine.  \secref{sec:related} discusses related work.  Finally, \secref{sec:conclusions} concludes the paper.

\section{Contextual-bandit Formulation}
\label{sec:bandit}


The contextual-bandit formalism~\cite{Barto85Pattern,Langford08Exploration} generalizes classic multi-armed bandits by introducing contextual information in the interaction loop between a \emph{learner} and the \emph{environment} it is situated in.  It has proved useful to model many important applications when such interaction is present, such as online advertising~\cite{Langford08Exploration} and content recommendation~\cite{Li10Contextual}.

Formally, we define by $\Aset=\{1,2,\ldots,K\}$ a set of \emph{actions}.  A contextual bandit describes a round-by-round interaction between a learner and the environment: at each round,
\begin{compactitem}
\item{Environment chooses contextual information $x\in\Xset$, and a \emph{reward} signal $r_a\in[0,1]$ for each action $a\in\Aset$, where $\Xset$ is the set of possible contextual information.  Let $\vec{r}=(r_1,\ldots,r_K)$ be the reward vector.  Only $x$ is revealed to the learner.  It is assumed that $(x,\hat{r})$ is drawn \iid\ from some unknown distribution $D$.}
\item{Upon observing $x$, the learner chooses an action $a\in\Aset$, and in return observes the corresponding reward $r_a$.}
\end{compactitem}
A common goal for a contextual-bandit learning algorithm is to optimize its action-selection policy, denoted $\pi:\Xset\rightarrow\Aset$, in order to maximize the expected reward it gets through interaction with the environment.  For convenience, we call
\[
V(\pi)\defeq\E_{(x,\vec{r})\sim D}\left[r_{\pi(x)}\right]
\]
the \emph{value} of policy $\pi$, which measures how much per-round reward it receives on average.  If we run $\pi$ to choose actions for $T$ rounds and observe the corresponding reward in every round, the value of $\pi$ can be estimated by averaging the observed reward, and this estimate converges almost surely to $V(\pi)$ as $T$ increases.

As an example, consider federated search where a search engine needs to decide, given a query, whether (and where) to include vertical search results like news and images on the final SERP (\eg, \cite{Diaz09Integration}).  Here, the context contains the submitted query, user profile, and possibly other information.  The actions are how to combine the vertical search result with Web search results.  The reward is often click-through (or its variants) of the vertical results.  Finally, a basic problem in federate search is to optimize the policy that decides what to do with vertical search results given current context in order to maximize the average reward.  In \secref{sec:exp}, we will study in more details another important component of a search engine.

An important observation in contextual bandits is that, only rewards of chosen actions are observed.  An online-learning algorithm must therefore find a good exploration/exploitation tradeoff, a defining challenge in bandit problems.  For offline policy evaluation, such partial observability raises a related difficulty.  Data in a contextual bandit is often in the form of $(x,a,r_a)$, where $a$ is the action chosen for context $x$ when collecting the data, and $r_a$ is the corresponding reward.  If this data is used to evaluate a policy $\pi$, which chooses a \textit{different} action $\pi(x)\ne a$, then we simply do not have the reward signal 
to evaluate the policy in that context.  It is the focus of this paper to study how unbiasded offline policy evaluation can be done when optimizing search engines.

\section{Unbiased Offline Evaluation}
\label{sec:eval}

\subsection{Basic Techniques}
\label{sec:eval-basic}

As is observed previously~\cite{Bottou13Counterfactual}, offline policy evaluation can be interpreted as a causal inference problem, an important research topic in statistics.  Here, we try to infer the average reward $V(\pi)$ (the causal effect) if policy $\pi$ is used to choose actions in the bandit problem (the intervention).  The approach we take in this paper relies on randomized data collection, which has been known to be a critical condition for drawing \emph{valid} causal conclusions~\cite{Rubin78Bayesian}.

Data collection proceeds as follows.  At each round,
\begin{compactitem}
\item{Environment chooses $(x,\vec{r})$ \iid\ from some unknown distribution $D$, and only reveals context $x$.}
\item{Based on $x$, one compute a multinomial distribution $\vec{p}\defeq(p_1,p_2,\ldots,p_K)$ over the actions $\Aset$.  A random action $a$ is drawn according to the distribution, and the corresponding reward $r_a$ and probability mass $p_a$ are logged.  Note that $\vec{p}$ may depend on $x$.  How to select the distribution $\vec{p}$ will be discussed later.}
\end{compactitem}
At the end of the process, we have a set $\Dset$ containing data of the form $(x,a,r_a,p_a)$.  We will call this kind of data ``exploration data,'' since all actions have some nonzero probability of being explored in the collection process.  In statistics, the probabilities $p_a$ are also known as \emph{propensity scores}.

When we are to evaluate the value of policy $\pi$ \emph{offline}, the following estimator is used
\begin{equation}
\Voff(\pi)\defeq\sum_{(x,a,r_a,p_a)\in\Dset}\frac{r_a \indfunc{\pi(x)=a}}{p_a} , \label{eqn:offline-estimate}
\end{equation}
where $\indfunc{C}$ is the set-indicator function that evaluates to $1$ if condition $C$ holds true, and $0$ otherwise.

The key observation of the estimator is that, for any context $x$, if one chooses action $a$ randomly according to the distribution $\vec{p}$, then
\[
r_{\pi(x)} = \E_{a\sim\vec{p}}\left[\frac{r_a\indfunc{\pi(x)=a}}{p_a}\right] .
\]
With this equality, one can show the unbiasedness of the offline estimator~\cite{Li10Contextual}: $\E_\Dset\left[\Voff(\pi)\right]=V(\pi)$ for any $\pi$, provided that every component in $\vec{p}$ is nonzero.  In other words, as long as we can randomize action selection, we can construct an unbiased estimate for any policy without even running it on users.  This benefit is highly desirable, since the offline evaluator allows one to simulate many A/B tests in a fast and inexpensive way.

\subsection{How to Randomize Data}
\label{sec:eval-data}

The unbiasedness guarantee holds for any probability distribution $\vec{p}$, as long as none of its component is zero.  However, \emph{variance} of our offline evaluator depends critically on this distribution.  The evaluator gives more accurate (and thus more reliable) estimates when variance is lower.

It follows from the definitions that the offline evaluator has a variance
\[
\Var\left[\Voff(\pi)\right] = \E_{(x,\vec{r})\sim D}\left[r_{\pi(x)}^2\left(\frac{1}{p_{\pi(x)}}-1\right)\right] .
\]
Therefore, the variance is smaller when we place more probability mass to actions that are chosen by policy $\pi$.  In reality, however, one typically does not know $\pi$ ahead of time when data are being collected, and there may be multiple policies to be evaluated on the same exploration data.  One natural choice, as is adopted by some authors~\cite{Li10Contextual,Li11Unbiased}, is to minimize the \emph{worst-case} variance, leading to a uniform distribution: $p_a\equiv1/K$ for all $a$.

There are at least two limitations for this choice.  First, choosing an action uniformly at random may be too risky for user experience, unless one knows \textit{a priori} that every action is reasonably good.  Second, when improving an existing policy that is already working reasonably well, it is likely that any improvement does not differ too much from it.  Minimizing the worst-case variance may not yield the best variance reduction in reality.  These two concerns imply a more conservative data collection procedure that can be more effective than the uniform random distribution.  Intuitively, given a baseline policy (like the existing policy in production), we may inject randomization to it to generate randomized actions that are close to the baseline policy.  The precise way of doing this depends on the problem at hand.  In \secref{sec:exp-data}, we describe a sensible approach that has worked well, which is expected to be useful in other scenarios.

Finally, in order to prevent unbounded variance, it is helpful to have propensity scores that are not too close to $0$, by ensuring a lower bound $\pmin>0$.  If such a condition cannot be met (say, due to system constraints), one can still use $\max\{\pmin,p_a\}$ to replace $p_a$ in \eqnref{eqn:offline-estimate}, as done by other authors~\cite{Strehl11Learning}.  Such a threshold trick may introduce a small bias to the offline estimator, but can drastically decrease its variance so that the overall mean squared error is reduced.

\subsection{How to Verify Propensity Scores}
\label{sec:eval-pscore}

As shown in \eqnref{eqn:offline-estimate}, it is necessary to compute and log the propensity scores $p_a$.  Any errors in the calculation and/or logging of the scores can lead to a bias in the final offline estimator.  Furthermore, since the \emph{reciprocal} of the score is used in the estimate, even a small error in the score can lead to a much larger bias in the offline estimator when $p_a$ is close to $0$.

When the system is complex, it is sometimes nontrivial to get the scores correct, even if a uniform distribution (\ie, $p_a\equiv1/K$) is used.  It is thus important to verify propensity scores before trusting the offline estimates.  In our experience, this verification turns out to be one of the most critical, and sometimes challenging, steps when doing offline evaluation.

One solution\footnote{Proposed by Leon Bottou \etal, in private communication.} is to obtain and log a randomization seed whenever an action is chosen.  Specifically, in each round of data collection (\eg, \secref{sec:eval-data}), we choose a seed $s$ (which may be a function of context $x$ and timestamp, \etc) and use it to reset the internal state of a pseudo-random number generator.  Then, we use the generator to select a random action from the multinomial distribution $\vec{p}$.  The final data have the form of $(x,s,a,\vec{p},r_a)$, containing more information to facilitate offline verification.  When we want to verify the propensity scores, we may simply use the seed $s$ to reproduce the randomized data collection process, and check consistency among $s$, $a$ and $\vec{p}$.

An alternative, somewhat simpler approach does not require resetting the pseudo-random number generator in every round of data collection.  Instead, it runs simple statistical tests to detect inconsistency.  Such an approach has been quite useful in our experience, although we note that it only detects some but not all data issues.

One such test, which we call an \textit{arithmetic mean test}, is to compare the number of times a particular action $a^*\in\Aset$ appears in the data to the expected number of occurrences conditioned on the logged propensity scores.  Concentration inequalities like Hoeffding's~\cite{Hoeffding63Probability} can be used to estimate whether the gap between the two quantities is statistically significant or not.  If the gap is significant, it indicates errors in the randomized data collection process.

Another test, which we found useful, is based on the following observation: for any context $x$ and action $a^*$,
\[
\E_{a\sim\vec{p}}\left[\frac{\indfunc{a=a^*}}{p_{a^*}}+\frac{\indfunc{a\ne a^*}}{1-p_{a^*}}\right] \equiv 2 .
\]
Therefore, we may compare the mean of the above random variable from the data, and verify if it is close to the expected value, $2$.  Again, statistical significance of the gap can be estimated by concentration inequalities.  Since the condition above uses harmonic means of propensity scores, it is called a \textit{harmonic mean test}.

\subsection{How to Construct Confidence Intervals}
\label{sec:eval-ci}

\eqnref{eqn:offline-estimate} gives a \emph{point}-estimate, which in itself is not very useful without considering variance information: when we compare the offline estimates of two policies' values, we must resort to reliable confidence intervals to infer whether the difference in the two point-estimates are significant or not.

Based on various concentration inequalities, Bottou~\etal~\cite{Bottou13Counterfactual} developed a series of very interesting confidence intervals.  The widths of the intervals can be used to gain helpful insights into the data collection process.  While the results are interesting, they are not necessarily the best candidate to use empirically, due to their worst-case nature.  As suggested by the same authors~\cite{Bottou13Counterfactual}, in reality it may be better to use normal approximation theory to obtain confidence intervals.  This is the approach we take in this work.

Specifically, from the exploration data set $\Dset$, we can compute an unbiased estimate of the standard deviation $\hat{\sigma}$ of the random variable
\[
\frac{r_a \indfunc{\pi(x)=a}}{p_a} .
\]
Then, a $95\%$ confidence interval can be constructed:
\[
\Voff \pm \frac{1.96\times\hat{\sigma}}{\sqrt{\setcard{\Dset}}} ,
\]
and so on.

\section{Case Study}
\label{sec:exp}

\subsection{Speller}
\label{sec:exp-speller}

Speller is a critical component for a search engine, enabling it to translate queries with typing and phonetic errors to their correct forms, so that it can match and rank relevant Web results and instant answers even when user-typed query is misspelt. Spelling correction for web queries is a hard problem, particularly because of absence of a dictionary of terms and new words and entities emerging on the Web as you are reading this. Further, one person's typo could be another person's correct query. For example, given that a user typed ``CCN'' has both the possibility of him wanting to type ``CNN'' but ending up making a typo to ``CCN,'' or really having an intent to type ``CCN.''  Typically, noisy channel models are applied to address this by computing probabilities of each of them being the true intents, given that ``CCN'' was typed, using popularities of ``CCN'' and ``CNN'' as well as how likely a user is to make this exact typo.  With additional features, machine learning can be used to rank these candidates~\cite{Gao10Large}.

The problem we focus on here is to train a model to select a subset of candidates off an already computed set of rewritten queries for a given input query. The idea to select multiple such candidates is to mitigate the risk of picking a bad correction early in the lifetime of the query~\cite{Sheldon11Lambdamerge}. After fetching the results for each of these formulations, we would either predict a single best rewrite or merge the results of multiple such rewrites. 

While training the ranker of rewritten query candidates via human-labeled corrections works for a large number of queries, in cases such as above, a judge may be at a complete loss on what the users' real intent was or predicting what is the likelihood of the two intents CCN and CNN. Hence, it is desirable to learn which spelling correction actually serves the user's intent implicitly from user behavior. Furthermore, this approach is much cheaper than a human judging the queries offline.  For a given spelling correction algorithm, the user's satisfaction can be measured by modeling how the user interacted with the search engine on a real search session.

Having said that, every single technique or an algorithm in the improvement of a search algorithm cannot be exposed to the user to measure its goodness (or badness), given that:
\begin{compactenum}
\item{It can pose risk to the relevance and quality of results seen by the users in the experiment.}
\item{User query volume restricts total number of experiments that can be run per unit time.}
\item{Cost of failure is higher online since typically more investment is required to make code robust enough to bring it in front of the users.}
\item{Online experimentation has more noise as well as harder to analyze the results of experiment against baseline, since the queries, users and sessions on which two algorithms were run differ.}
\end{compactenum}
The concerns above lead to a need of an offline evaluation system even when the labels at first place have been collected via online data.


Using terminology in \secref{sec:bandit}, the context includes the user-typed query and rewritten candidates (together with their features); an action is to decide which candidate(s) to use;
and the reward is a metric derived from user clicks on the final search result page.  Due to business sensitivity, the metric is not revealed, although it suffices to say that the metric measures goodness of the final SERP for the present user.  From now on, this metric is referred to as the \emph{target metric}.   Apparently, the actions affect the final SERP, which in turn affect user clicks and the reward. 


\subsection{Data Collection}
\label{sec:exp-data}

Our data collection process was run on a small fraction of random users (identified by browser cookies) for a week in late 2013, yielding about 15M impressions.  This is the exploration data set $\Dset$ to be used in the following experiments.

To avoid adverse user experience, we require the top candidate must be included in the action for every impression.  Other candidates were sent randomly: for $i\in\{2,\ldots,L\}$,
\[
\Pr\left(q_i\mbox{ is chosen}\right) \defeq \frac{1}{1+\exp(\lambda_1(s_1-s_i)+\lambda_2)}
\]
where $\lambda_1$ and $\lambda_2$ were parameters tuned to yield a good balance between aggressiveness of exploration and potential negative user experience.  In our case, the parameters were chosen so that the path-selection probabilities fell in $[0.1,0.9]$, and that an offline metric based on judge labels was not severely affected.
With these probabilities, propensity scores of actions can be computed.

There are a few benefits of using this randomization scheme.  First, by always including $q_1$ in $P$, the final SERP is usually reasonably good, so users who were included in the data-collection process would not notice much decrease in relevance quality.  Second, the scheme is motivated by the intuition that candidates with higher scores tend to be better, so are more likely to be chosen by a good policy.  Biasing data collection towards such more promising candidates will likely reduce variance, following the discussion in \secref{sec:eval-data}.

After data were collected, we ran the arithmetic and harmonic mean tests described in \secref{sec:eval-pscore} and found no major issues.  It is worth mentioning that the tests were able to help detect data issues in earlier versions of data collection, leading to fixes that eventually improved data quality.

\subsection{Bootstrapping Histogram}
\label{sec:exp-bootstrap}

In \secref{sec:eval-ci}, we advocate the use of asymptotic normal approximation to construct confidence intervals.  The underlying assumption is that, when the amount of data is large, the estimator in \eqnref{eqn:offline-estimate} is almost normally distributed.  Here, we verify this assumption empirically with bootstrapping.

In particular, we sampled impressions from $\Dset$ with replacement to get a bootstrapping set $\Dset_1$ of the same size as $\Dset$.  The sampling procedure was repeated $B=1000$ times, and we ended up with $B$ data sets: $\Dset_1,\Dset_2,\ldots,\Dset_B$.  On each of them, we computed the unbiased offline estimate of click-through rate (CTR) for some fixed policy $\pi$.  Finally, we had $B$ estimates of CTR for the same policy.  Since we were dealing with a large data set, we implemented the \emph{online} bootstrap~\cite{Oza01Online}, which is essentially identical to the standard bootstrap for the size of our data.

\begin{figure}
\begin{center}
\includegraphics[width=0.8\columnwidth]{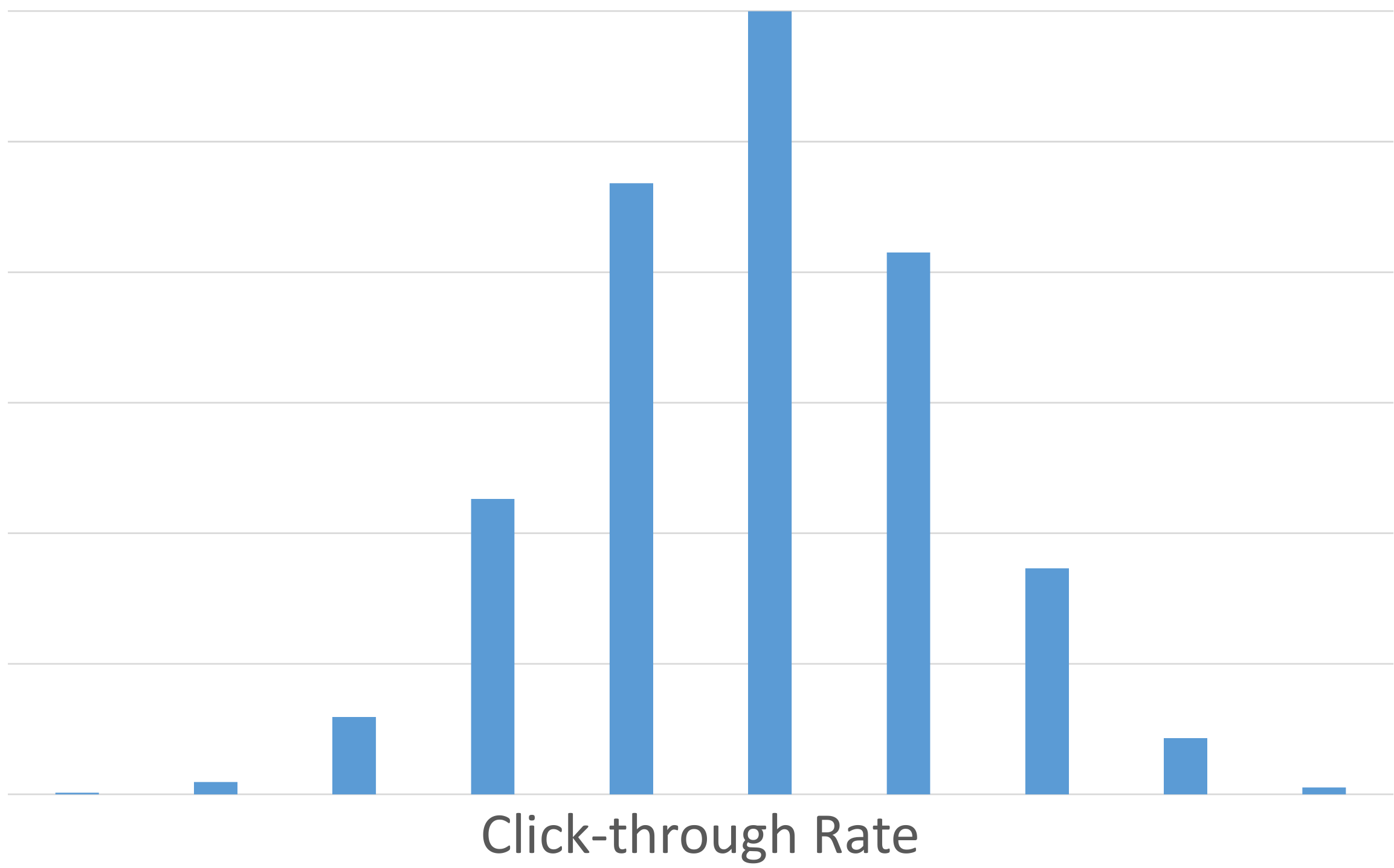}
\end{center}
\caption{Bootstrapping histogram.} \label{fig:bootstrap}
\end{figure}

\figref{fig:bootstrap} gives the histogram of the $B$ estimates of CTR.\footnote{All CTRs and target metrics reported in the paper are normalized (\ie, multiplied by a constant) for confidentiality reasons.}  It is rather close to a normal distribution, as expected.  The result thus validates the form of confidence intervals given in \secref{sec:eval-ci}.  Given the size of our data, the confidence intervals computed this way were all tiny, usually on the order of $10^{-2}$ or less.  We therefore did not include these tiny intervals in the plots.

\subsection{Accuracy against Ground Truths}
\label{sec:exp-accuracy}

We now investigate accuracy of the offline evaluator.  During the same period of time when $\Dset$ was collected, we also ran another candidate selection policy $\pi'$.  The online statistics of $\pi'$ could then be used as ``ground truth,'' which can be used to validate accuracy of the offline estimator using exploration data $\Dset$.

\begin{figure}
\begin{center}
\includegraphics[width=\columnwidth]{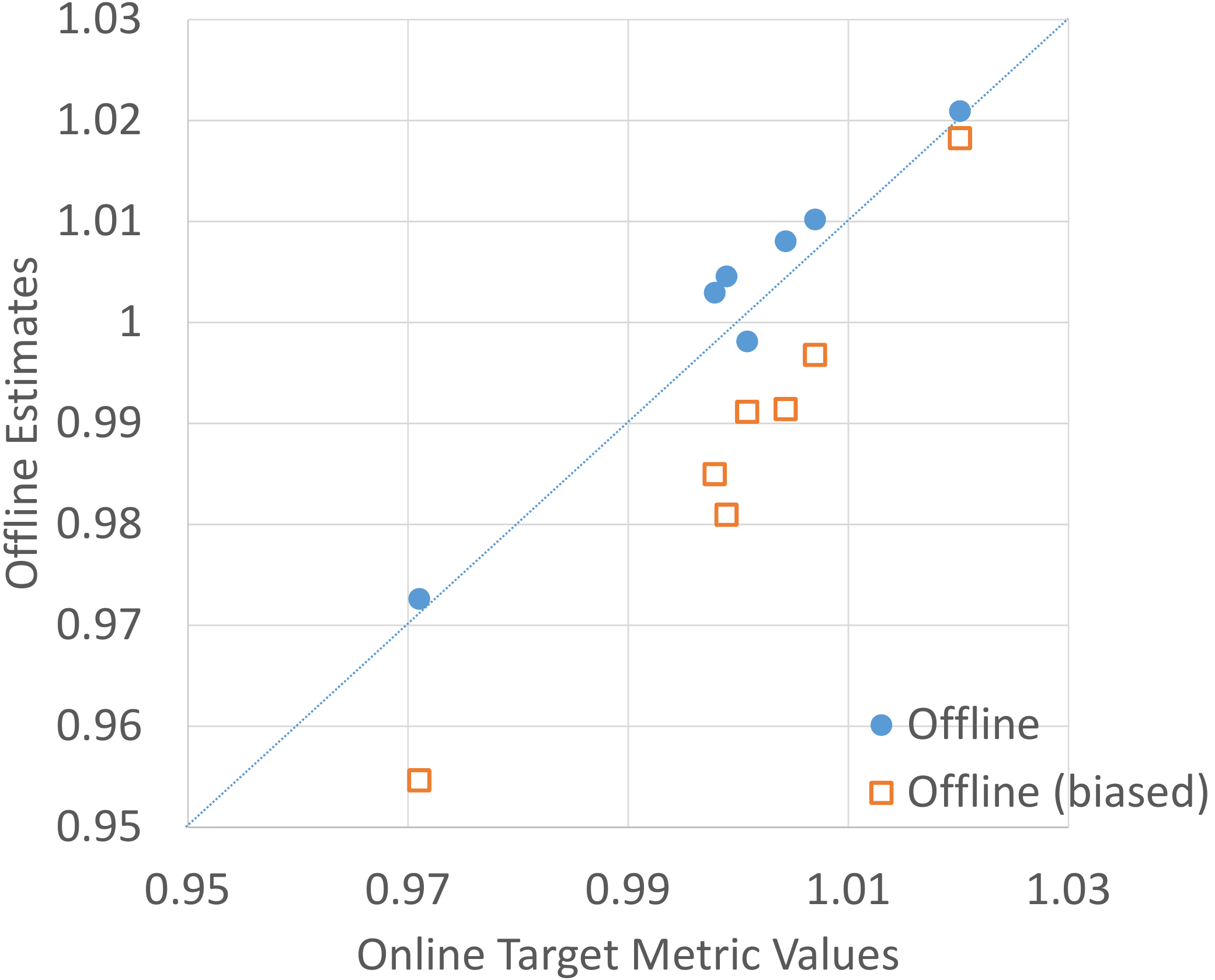}
\end{center}
\caption{Scatter plot of the online vs. offline metric values.  Each point corresponds to one of the seven days in the data collection period.} \label{fig:pcr}
\end{figure}

First, we examine the estimate of the target metric for each day of the week.  \figref{fig:pcr} is a scatter plot of the online (true) vs. offline (estimated) values.  As expected, the offline estimates are highly accurate, centering around the online ground truth values.  Also included in the plot is a biased version of the offline estimate, labeled ``Offline (biased),'' which uses the following variant of \eqnref{eqn:offline-estimate}:
\[
\Voff^{\mathrm{(biased)}}(\pi)\defeq\frac{\sum_{(x,a,r_a,p_a)\in\Dset}r_a\indfunc{\pi(x)=a}}{\sum_{(x,a,r_a,p_a)\in\Dset}\indfunc{\pi(x)=a}} .
\]
This estimator, which is not uncommon in practice, ignores sampling bias in $\Dset$.  The effect can be seen from the much larger estimation error compared to the online ground truths, confirming the need for using the reciprocal of propensity scores in \eqnref{eqn:offline-estimate}.

\begin{figure}[h]
\begin{center}
\includegraphics[width=\columnwidth]{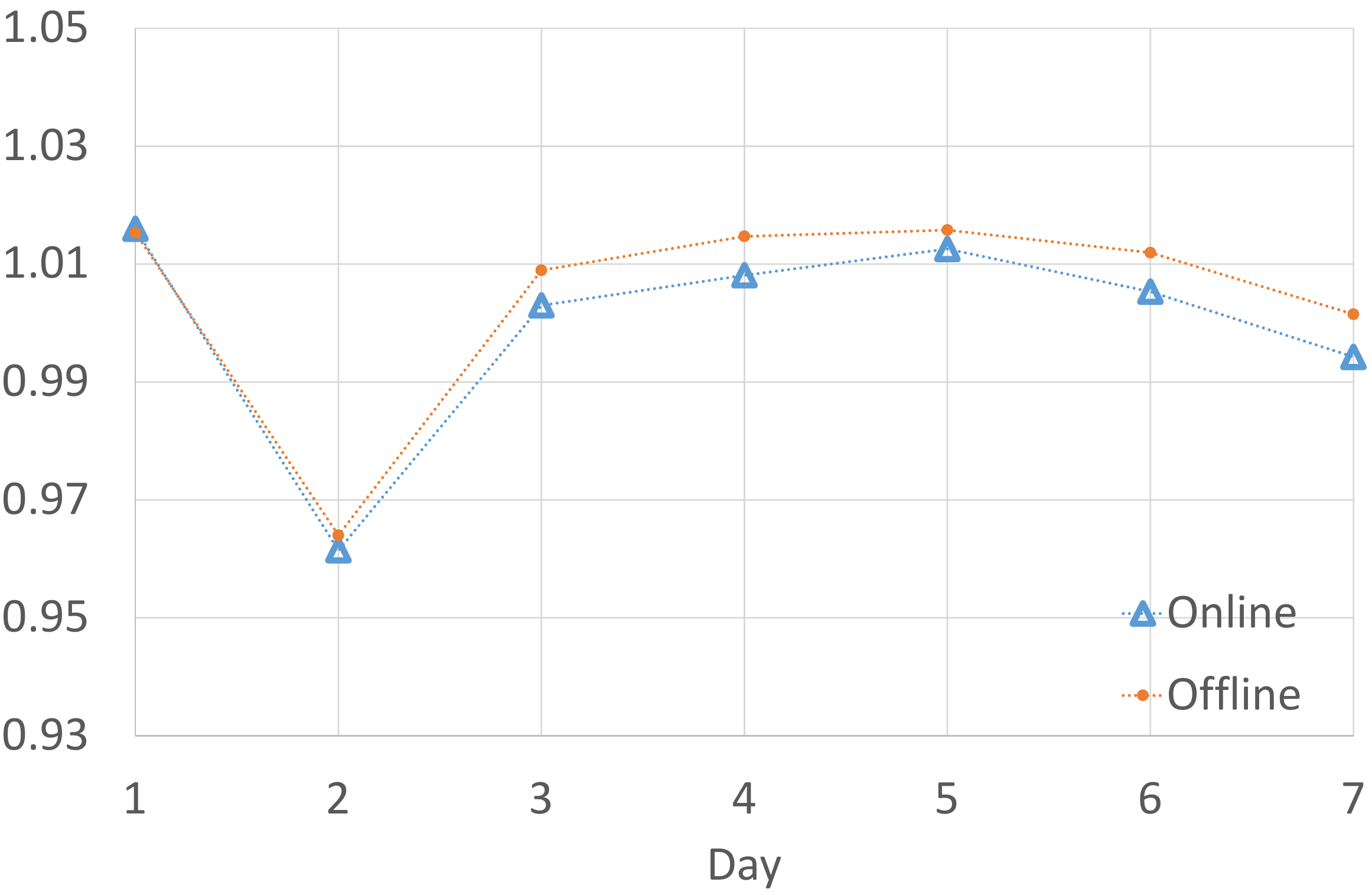}
\end{center}
\caption{Daily CTR.} \label{fig:ctr-day}
\end{figure}

\begin{figure}
\begin{center}
\includegraphics[width=\columnwidth]{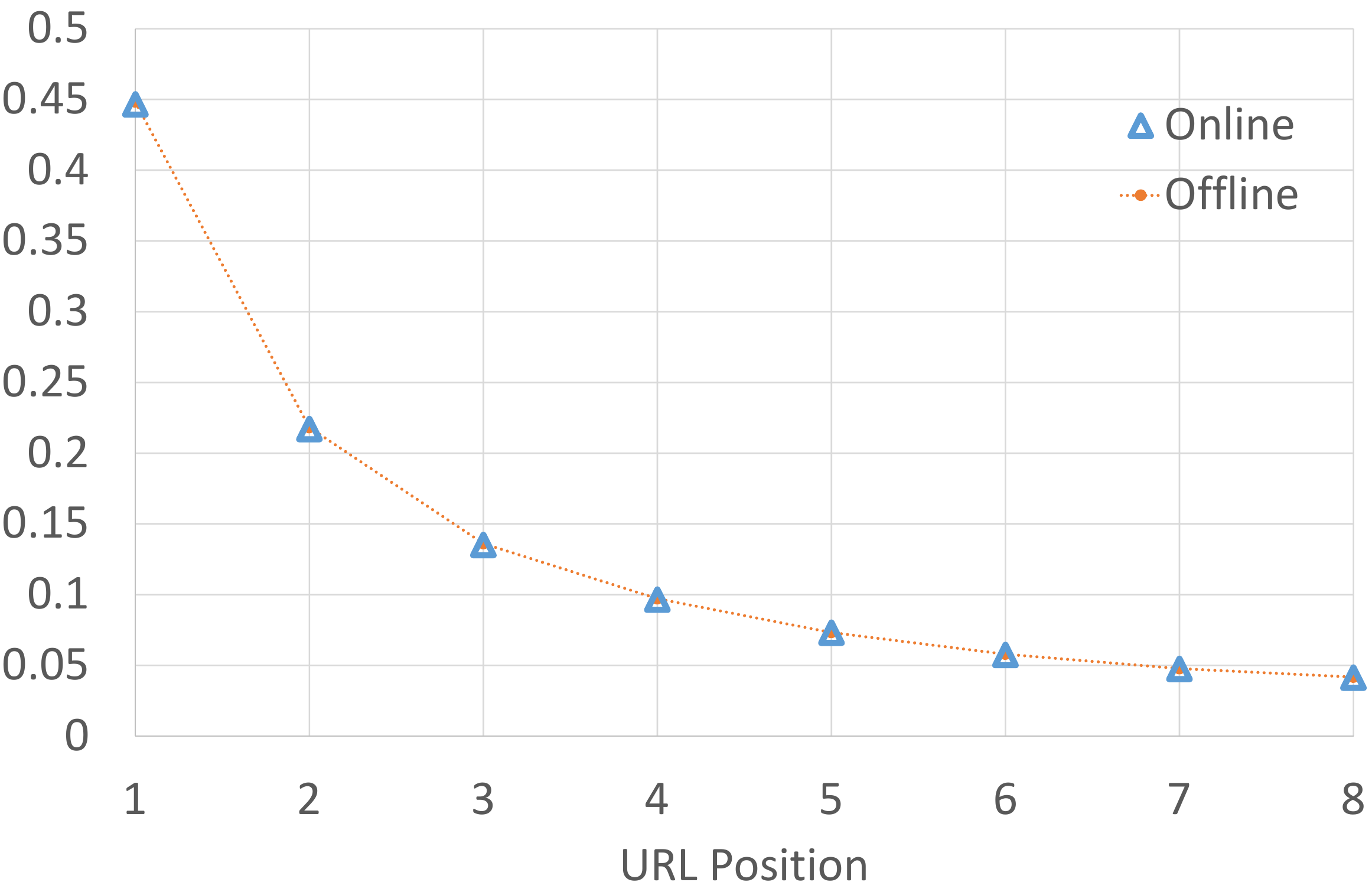}
\end{center}
\caption{CTR as a function of position on SERP.} \label{fig:ctr-pos}
\end{figure}

\begin{figure}[h]
\begin{center}
\includegraphics[width=\columnwidth]{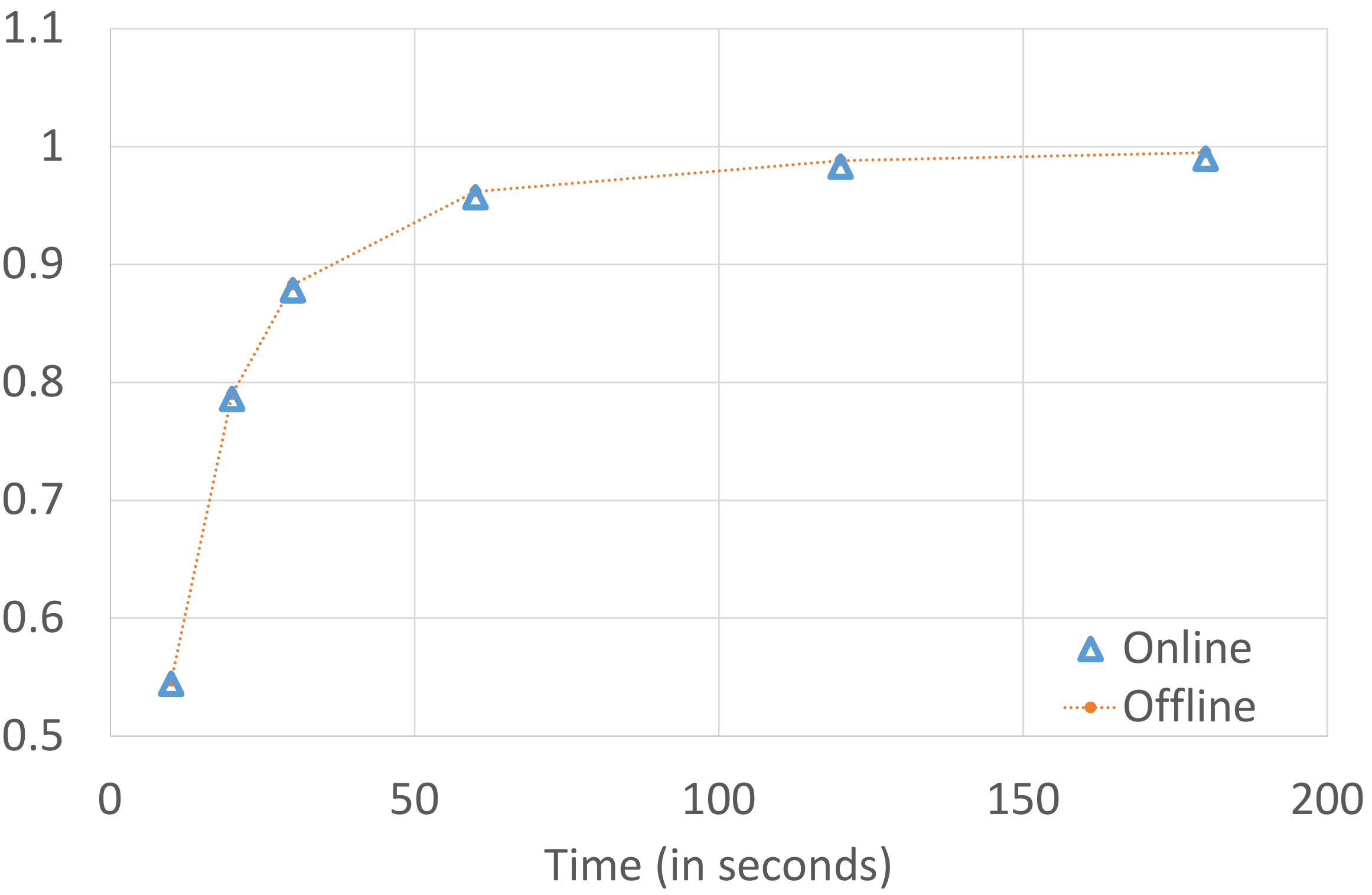}
\end{center}
\caption{CTR as a function of time (in seconds).} \label{fig:ctr-time}
\end{figure}

We now look at the CTR metric more closely.  \figref{fig:ctr-day} plots how daily CTR varies within one week.  Again, the offline estimates match the online values very accurately.  The gap between the two curves is not statistically significant, as the $95\%$ confidence intervals' widths are roughly $0.01$.  \figref{fig:ctr-pos} gives the fraction of clicks contributed by URLs of different positions on the SERP.  \figref{fig:ctr-time} measures how many clicks were received within a given time after a user submitted the query.  All results show high accuracy of the offline evaluator.  We have also done the same comparison on other metrics and observed similar results.

\subsection{Offline Optimization}
\label{sec:exp-optimization}

While offline evaluation can be very useful on its own, reducing a substantial fraction of online A/B testing, it can be used as a subroutine for offline \emph{optimization} of policies.

We used a different data set similarly collected from fall 2013, and split it randomly into a training set ($2/3$) and an evaluation set ($1/3$).  From the training set, we extracted training data whose (binary) labels were determined based on whether the sent candidate contributed positively to the target metric.  This way, we obtained a binary classification problem, in which one tries to predict whether a rewritten query will contribute to the target metric, conditioned on the original query.  Boosted logistic regression was used to learn a model.

When learning such a model, a few meta-parameters had to be chosen to balance the target metric improvement and capacity constraints.  One could have used prediction accuracy on a hold-out set to select them, but accuracy does not necessarily correlated with the target metric that we aim to optimize eventually.  Fortunately, the reliable offline evaluator can be used to select these parameters to optimize the target metric \emph{directly} while respecting capacity constraints.  Based on the offline evaluation results, we picked a model and ran an A/B test.  In a two-week experiment done in late 2013, the model did show a statistically significant improvement over the existing baseline, demonstrating the power of the offline evaluator.  Below, we demonstrate a few successful cases where the new model improves upon the baseline.

In the first example, the user-typed query was ``umecka and zinc.''  Our policy successfully identified the corrected query ``umcka and zinc,'' which is about medical treatments for cold symptoms relief.  The SERP included an Amazon page about a product (Umcka Cold and Flu Chewable) and customer reviews comparing to zinc supplements (another common treatment for similar purposes).  Analyzing the user clicks on the SERP suggested the user found the needed information.  In contrast, the baseline as well as other similar commercial search engines appeared to miss the correction, and only showed results that contained an exact match to ``umecka.''

In the second example, the original query submitted by user was ``catalina left attorney.''  The new policy correctly suggested ``catalina leff attorney,'' which appeared to be the right correction: there is an attorney at San Diego whose name is Catalina Leff.  The baseline failed to identify this correction, and only show results that contains ``left,'' which was probably not what the user really intended.

\section{Related Work}
\label{sec:related}

There is a long history of evaluation methodology research in the information retrieval community~\cite{Robertson08History}.  The dominant approach is to collect relevance judgments for a collection of query-document pairs, and compare different retrieval/ranking functions against metrics like mean average precision~\cite{Baezayates99Modern} and normalized discounted cumulative gains~\cite{Jarvelin02Cumulated}.  This approach has been very successful as a low-cost evaluation scheme.  However, several authors have argued for several of its limitations (\eg, \cite{Belkin08Some,Turpin01Why}), in addition to the ones discussed in \secsref{sec:intro} and \ref{sec:exp-speller}; an alternative, ``user-centered'' evaluation emphasizes interaction between user and the search engine.  One challenge with the user-centered approach is the relatively high cost for system evaluation and comparison.  Our work, therefore, provides a promising solution that has shown success in an actual search engine.

In industry, people have also measured various online metrics (such as CTR and time to first click) to monitor and compare systems while running them to serve users.  Randomized control experiments (\eg, \cite{Kohavi09Controlled}) are the standard way to measure and compare such online metrics.  More recently, interleaving has become an attractive technique to quickly identify the winner when comping two systems~\cite{Chapelle12Large}.  Both techniques requires running a system on real users, while our offline approach here can be more efficient and less expensive.

As mentioned earlier, the offline evaluation technique is closely related to causal inference in statistics~\cite{Holland86Statistics}, in which one aims to infer, from observational data, the counterfactual effect on some measurement by changing the policy (more often called ``intervention'' in the statistics literature).  Such counterfactual methods have shown promise in a few important Web applications recently like advertising~\cite{Bottou13Counterfactual,Chan10Evaluating,Lambert07More,Langford08Exploration,Tang13Automatic}, content recommendation~\cite{Li11Unbiased}.  In this work, we formulate the problem in the contextual bandit framework as in \cite{Langford08Exploration,Li11Unbiased},
which is natural to model such interactive machine learning problems.  Furthermore, although offline evaluation was applied to recency search in the past~\cite{Moon12Online}, to the best of our knowledge, this work is the first to demonstrate effectiveness of counterfactual analytic techniques for Web search, including head-to-head comparisons and offline optimization in a commercial search engine.

\section{Conclusions}
\label{sec:conclusions}

In this work, we formulate a class of optimization problems in search engines as a contextual bandit problem, and focus on the \emph{offline} policy evaluation problem.  Our approach uses counterfactual analytic techniques to obtain an unbiased estimate of the true policy value, without the need to run the policy on real users.  Using data collected from a commercial search engine, we verified the reliability of such an evaluation, and also showed a successful application of it for offline policy evaluation.

The promising results in this paper suggest a number of interesting directions for future research.  The action set in Speller is tractably small when we only consider a short list of candidates.  The set of actions in a ranking problem, defined naively, consists of all permutations of URLs.  This is an exponentially large set that can cause the variance to be large.  It would be interesting to see how to leverage successful ideas in related work~\cite{Bottou13Counterfactual,Dudik11Doubly} to address this issue.  Another direction worth investigating is direct optimization of policies based on exploration data with, for instance, the offset tree algorithm~\cite{Beygelzimer09Offset}.

\subsection*{Acknowledgements}

This work benefited from helpful discussions with Leon Bottou, Chris Burges, Nick Craswell, Susan Dumais, Jianfeng Gao, John Platt, Ryen White, and Yinzhe Yu.

\bibliographystyle{plain}
\bibliography{refs}

\balancecolumns

\end{document}